\documentclass[letterpaper]{article}
\usepackage{aaai}
\usepackage{times}
\usepackage{helvet}
\usepackage{courier}
\usepackage{dcolumn}
\usepackage{color}
\usepackage{fleqn}
\usepackage{algpseudocode, algorithm}
\usepackage{amsmath,amssymb}
\usepackage{graphicx}
\newtheorem{definition}{Definition}[section]

\newtheorem{prop}{Proposition}[section]

\newcommand{\cout}[1]{}

\begin{document}
%
\title{Semantic Preserving Generative Adversarial Models}
\cout{
\author{Shahar Harel \thanks{SparkBeyond LTD, email: shahar@sparkbeyond.com} \And Meir Maor\thanks{SparkBeyond LTD, email: meir@sparkbeyond.com} \And Amir Ronen\thanks{SparkBeyond LTD, email: amir.ronen@sparkbeyond.com}\footnote{Equal contribution paper}}}

\author{Shahar Harel \thanks{Email: shahar@sparkbeyond.com}, Meir Maor \thanks{Email: meir@sparkbeyond.com}, Amir Ronen \thanks{Email: amir.ronen@sparkbeyond.com}\\
SparkBeyond LTD\\
Israel\\
}

\maketitle

\begin{abstract}
We introduce generative adversarial models in which the discriminator is replaced by a calibrated (non-differentiable) classifier repeatedly enhanced by domain relevant features. The role of the classifier is to prove that the actual and generated data differ over a controlled semantic space. We demonstrate that such models have the ability to generate objects with strong guarantees on their properties in a wide range of domains. They require less data than ordinary GANs, provide natural stopping conditions, uncover important properties of the data, and enhance transfer learning. Our techniques can be combined with standard generative models. We demonstrate the usefulness of our approach by applying it to several unrelated domains: generating good locations for cellular antennae, molecule generation preserving key chemical properties, and generating and extrapolating lines from very few data points. Intriguing open problems are presented as well.
\end{abstract}

\section{Introduction}\label{sec:intro}
Generative adversarial networks (GANs) \cite{gansGoodfellow} achieved many impressive results. Recent literature surveys as well as a large code repository can be found at \cite{ganOverview,ganLandscape,GanZoo}. Arguably however, most of these results were obtained for generation of images, text, and videos. These domains exhibit several special properties that aid in their success. First, humans have very good judgment of the quality of the generated objects and hence can fine-tune the generative model until it is satisfactory. Second, there exists a huge amount of available data that can be used for model training. Third, the body of knowledge available in these domains facilitates highly effective representations, utilization of existing topologies, etc. This is unlikely to be the case in a wide range of important domains (e.g. generating trajectories of vehicles or traces of sensory data, producing artificial health records \cite{generatingHealthRecords}, creating plans for 3D-object printing). In many domains it will be hard to assess the quality of the results. Instead we would like to compare the generated and actual data across a range of domain specific properties. Furthermore, it is often the case that large amounts of data are difficult to obtain. 

Consider for example, the discovery of novel chemical formulas that can be used for the production of a fertilizer. We would like the molecule to be stable from a chemical perspective, to have a reasonable production cost, to have toxicity profile similar to existing fertilizers, etc. In conventional GANs, even if the generator and the discriminator reach equilibrium, the fact the discriminator neural network fails to separate the generated and the actual data does not guarantee such compound properties. Moreover, there may be properties of existing fertilizers that we are not aware of that should also be considered. Ideally, we would like to construct models that will first establish deep understanding of the common  properties of the data at hand, and  then generate objects with similar properties. Desirably, we would like this model to be transparent about its properties so the user will be able to decide which properties are important to preserve and which can differ. 

In this paper we propose a generic method of constructing generative models that preserve semantic properties of the actual data.  We assume the existence of a set of {\em semantic functions} denoted by $\cal{F}$ where each $f \in \cal{F}$ is a computable function $f: \cal{X} \rightarrow R$. $\cal{F}$ is given as {\em input} to our framework and represents the properties that the user cares about (e.g. kinetic properties of trajectories, sentiment or literacy level of text, demographics of locations). The functions in $\cal{F}$ are provided as black boxes and are not necessarily differentiable or even continuous. Our high level goal is to produce objects that are indistinguishable from the actual data given a natural space of predicates spanned by the semantic functions. One should recognize that this typically creates a large space of natural criteria that the generated data should meet. In many reasonable settings this space might be much larger than the size of the available data triggering what is often called the $p \gg n$ problem in statistics \cite{Hastie03expressionarrays}. The proposed method is based on the classic GAN framework \cite{gansGoodfellow} with several fundamental differences outlined in Section \ref{sec:semanticGans}. In a nutshell the discriminator is replaced by a component termed semantic engine whose goal is to ``prove'' that the actual and generated data have different semantic properties. Once the engine fails we get strong guarantees on the generated data. We outline the ideal properties of such component and suggest how to approximate them in practice. This yields natural stopping conditions and control over the generator-discriminator co-training. We use the REINFORCE trick \cite{reinforce} to update the generator. Our method is first introduced in a discrete set-up. We then show how to scale it to sequence generation. For sequences we also suggest a method of reducing mode-collapse via extrapolation tasks. We then modify our method to handle various types of continuous data. All variants are applied in experiments. 

The experimental part is composed of three problems of very different characteristics. 
We start with a problem that appears different than traditional problems to which GANs were applied in the past. Specifically, we strive to automatically characterize and generate good locations for cellular antennae given only a few hundreds of training points (latitude-longitude pairs). We demonstrate that later generations of the method indeed yield generated data that is closer to the actual data in a large semantic space. We use the generated data in order to transfer the results to other areas and to identify insightful anomalies. 

In order to facilitate comparison to state of the art methods we conduct experiments in molecule generation. We conduct experiments in which we are given a source family of molecules and the objective is to generate novel molecules in a way that preserves the distributions of key chemical properties. Recently, \cite{GuacaMol} showed near perfect results when the family of molecules is very large (over a Million). We show that these results significantly deteriorate with the number of training data points and that our method can significantly improve the results in this case.

Finally, in order to systematically explore some properties of our method, we study an artificial problem of generating straight lines given only a handful of training examples.  

We believe that the method proposed here can help applying GANs to many novel domains. It allows to get essential guarantees on the results and to get a clear signal when things go wrong. The usage of semantic functions introduces prior knowledge into the data. As demonstrated, this allows generalization from a smaller number of points. This work gives rise to several intriguing open issues. We outline them in the concluding section.

\subsection{Related work}
The basic idea of GANs was introduced in \cite{gansGoodfellow} and was further explored by a large body of work. The properties of the resulting Nash-Equilibrium along with the generalization power of GANs in a high dimensional space were explore, e.g by \cite{AroraGanBounds,thanh-tung2018improving}. Variants such as Wasserstein \cite{wassersteinGans} or MMD \cite{mmdGans} GANs can be viewed as attempts to overcome the above shortcomings. Tuning them for a high dimensional set of non-differential features appears challenging though. Some literature on semantics of generated objects was considered in the past, (e.g. \cite{semanticImageInpainting,insepctionDistance}). However, these works deal with differentiable semantics and are designed specifically in the context of image generation.

Generating new molecules and materials is an area of great interest recently (e.g. \cite{gomez2018automatic,shahar,molecularNature}). We use this domain to compare our work with state of the art. For this comparison we use the evaluation framework of \cite{GuacaMol}. We demonstrate the importance of our method when the amount of available data becomes smaller. Note that some of the work on molecule generation is either goal oriented (e.g. \cite{GoalDirectedMolecularGraph,Molecular-de-novo-design,Organ}), use molecule representations other than ours, or based on auto-encoders \cite{ConstrainedGraphVariationalAutoencodersMolecule,VariationalAutoencoderMolecularGraph}. While combining such methods with ours appears possible we do not see merit in comparing them directly to our work.
Some of the above work involves reinforcement learning in order to optimize specific goals. Our work, in contrast, aims to preserve spaces of semantic properties whose exact shape might not even be known in advance.

\section{Semantic preserving GANs} \label{sec:semanticGans}
This section presents the main ideas behind our method. We first present some key concepts. Our construction is introduced at Section \ref{sec:basicGans} and various extensions are presented afterwards. 

The classic GAN framework \cite{gansGoodfellow} is a zero sum game between two differentiable neural networks, a discriminator and a generator, where the goal of the discriminator is to distinguish between the actual data and the generated data, and the goal of the generator is to fool the discriminator. We follow this approach. We introduce the following key modifications. 

\begin{enumerate}
    \item Feature generation and selection component: Given the set of semantic functions $\cal{F}$, this component generates a set of features $h:{\cal{X}} \rightarrow \{ 0 ,1 \} $ over a well defined family based on $\cal{F}$ . For example $f(x) > \alpha$ or $f(x)/g(x) < 0$. In order to prevent over-fitting the component chooses a small subset of the features with strong combined predictive power. This component should correct for the size of the space of potential features. The actual and generated data are enriched with the set of selected features.
    
    \item We use a calibrated classifier with strong over-fitting prevention policy (e.g. a logistic classifier, xgboost with aggressive stopping rules) to discriminate between the actual data and the generated data. 
    
    \item We maximize an expected reward over the generated data. The reward is given by the probability to fool the discriminator (i.e. the discriminator predicted probability of generated data being real).
    We use the REINFORCE trick \cite{reinforce} in order to update the gradients of the generator.
    
    \item We retrain the discriminator once the average reward gets close to 0.5 or once a certain number of iterations have passed. We use the validation set's area under curve (AUC) as a stopping criterion.
\end{enumerate}
A good metaphor for our framework is a {\bf theorem proving} game. 
The discriminator constantly attempts to generate proofs
for statistical differences between the underlying distributions of the generated and the actual data. An AUC level near 0.5 means that the discriminator is not able to show that the two distributions differ. This gives a strong {\bf certificate} on the quality of the generated objects. For this schema to work we need to construct a discriminator with properties that we discuss in the sequel. 

If the process is successful, it ends up with sufficiently small AUC. If the AUC remains far from 0.5 the user learns something is wrong (e.g. due to insufficient data, a weak generator, or constant mode switching). The user can also learn a lot from the nature of the separating features.

\subsection{Semantic Engines}
It is widely accepted that, in many cases of interest, feature engineering is imperative for building robust machine learning models and allows generalization from fewer samples. In its manual form, feature engineering lets humans inject their own understanding of the world which often goes well beyond the data-set. 

We term {\em semantic engine} a component which leverages a collection of functions ${\cal F}$, each maps samples from the input space ${\cal X}$ to $R$, representing some meaningful property. Given a binary labelled dataset the engine constructs a classifier that separates between the label classes using features based on ${\cal F}$. 

A naive semantic engine may simply enrich the raw data with the engineered features and then build a standard machine learning model on the enriched data. Such a simple approach is sufficient for some of our experiments. A more sophisticated engine will apply a feature selection algorithm and calibration procedures. There are more advanced commercial and open-source solutions, which build combinatorial combinations to span a very large potential feature space and employ aggressive feature selection and model build strategies to prevent over-fitting. Obviously, curating a good collection of potential features is key to building a successful semantic engine and feature selection strategies are heuristic which come with few guarantees. Nonetheless empirical evidence suggests that such an approach can build accurate generalizable models. 

Another practical advantage of both semantic engines and engineered features is their transparency. For example, a feature like "The average current is above 5 Ampere" is easily understandable by a human. This gives clarity regarding meaningful differences between generated and real data (if exist) and about what is being optimized by the algorithm at each stage. 

\subsection{Ideal properties of semantic engines}
It it helpful to consider what are the ideal properties of a classifier generated by the semantic engine. We will then use these properties to both guide the co-training of the generator and discriminator and as a stopping condition. 

\begin{definition}\label{def:idealDisc}{\bf (ideal discriminator)} Let $D$ be a distribution over the input space $\cal{X}$. Let $\cal{H}$ be a set of potential features of the form $h:{\cal{X}} \rightarrow \{ 0, 1 \}$. An {\em ideal discriminator} for D over $\cal{H}$ is a function $e:{\cal{X}} \rightarrow [0, 1]$ such that there exist two constants $0 < \alpha < 1$ and $c > 0$ such that for every distribution $G$ over $\cal{X}$:
\begin{description}
\item[separability] If there exists a potential feature $h \in \cal{H}$ such that $|E_{x \in D}[h(x)] - E_{z \in G}[h(z)]| > c \cdot \alpha$ than $e$ separates $D$ from $G$ with $AUC > 1/2 + \alpha$. That is: $Pr_{d \in D, g \in G} \, [e(d) > e(g)] +  \frac{1}{2} \cdot Pr_{d \in D, g \in G} \, [e(d) = e(g)] > \alpha$\footnote{AUC is a shorthand of the ROC AUC statistic. The above probabilistic interpretation is known to be equivalent to the more common definition via an  integral.}.
\item[properness] For all $G$, $AUC(e) \geq 1/2$
\end{description}
\end{definition}
In other words, $D$ represents the actual distribution. The discriminator returns a probability of an instance being actual. The discriminator learns to distinguish between $D$ and $G$ such that: (a) the existence of a separating feature guarantees a separation by the discriminator and (b) the discriminator avoids overfitting. The space $H$ is based on natural hypotheses over ${\cal F}$ as described above. While, for finite samples, under a substantial space of potential features, we cannot always expect the existence of such a discriminator, it can be approximated in many practical settings.

The following key property immediately follows. 

\begin{prop} \label{def:keyProperty} {\bf (key property)} Under the conditions of Definition \ref{def:idealDisc}, there exists a constant $0 < \gamma < 1$ such that if for $D$ and $G$ $AUC(e) < 1/2+ \alpha$, then for every potential feature $h \in H$, $|E_{x \in D}[h(x)] - E_{z \in G}[h(z)]| \leq \gamma \cdot \alpha$  
\end{prop}
The key property enables the definition of a natural {\bf stopping} condition as well as a criterion to when to re-train the discriminator. Once the AUC becomes low it is guaranteed that {\em all} potential features in $\cal{H}$ have expected values which are similar on the actual and the generated data. On the other hand, if the AUC is not improving, the user knows that the system is not progressing well. 
In order to approximate the two properties in practice we use an aggressive non-overfitting strategy. While we don't explicitly calibrate the classifier in our experiments (e.g. via isotonic regression) we focus on classifiers whose goal is to issue calibrated predictions such as logistic regression or gradient boosting with relatively small number of features.

\subsection{Basic semantic preserving GAN}
\label{sec:basicGans}
We now introduce our basic construction, termed semantic preserving GAN (SPGAN), focusing on a discrete setup. In the next subsections we propose an adaptation to continuous setup and introduce a scalable generalization to sequences. The GAN is presented in Algorithm \ref{alg:gan1}. It is somewhat simplified to make it more readable. We assume a discrete set of outputs $B = 1 \ldots b$. The GAN always outputs a {\em probability} over $B$ and improves it iteratively until the AUC drops to a sufficient level or the maximal number of iterations allowed has been reached. This stopping condition is justified by the key property. At each iteration the semantic engine constructs a new discriminator according to recent generated data (mixed with the previous generation for smoothness) and the actual data. Each iteration is divided into batches on which we update the generator.

More formally, the generator maximizes an expected reward defined below. We use the REINFORCE \cite{reinforce} trick to derive the gradients for the reward function 
\begin{equation}
\begin{split}
    J(\theta) = \mathop{\mathbb{E}}[R|\theta] = \sum_{j \in outputs} p_j \times l_j \: \: \: \: \: \: \: \: \: 
    \nabla J(\theta) \\
    = 
    \sum_{j \in outputs} p_j \times \nabla_{\theta}(\log(l_j))
\end{split}
\end{equation}

where $l_j$ is the generator probability for output $j$ being generated and $p_j$ is the reward given by the probability that the discriminator assigns for this output of being real.  $\theta$ is the generator's parameter vector. Intuitively, the generator attempts to fool the discriminator by maximizing the probabilities the discriminator assigns to the generated data being real.  
Once the average probability of a batch approaches $0.5$ (or a sufficient number of iterations have passed) we end the iteration. 
On most of our experiments the generator architecture is very simple. It is based on a shallow feed forward network or on an LSTM for sequence generation. A soft-max operation over the final layer returns a probability over the discrete set of outputs.  

\begin{algorithm}[h]
\hspace*{\algorithmicindent} \textbf{Main Input} Actual data, Semantic engine $e$, threshold  $\epsilon$, output space $B$, generator network with parameter set $\theta$ \\
\hspace*{\algorithmicindent} \textbf{Output} Generator  and sample generated data
\begin{algorithmic}[1]
\While  {$AUC > 0.5+ \epsilon$} 
    \State Produce train data from two equal sized sets. Positives: samples from the actual data. Negatives: a mix of newly generated samples and the generated samples of the previous training phase. 
    \State Retrain the semantic engine $e(.)$ on the above to get a calibrated classifier and out of sample AUC estimation
	\While {$batch \_ mean \_ score < 1/2 - \epsilon$ or sufficient number of steps}
	    \State Generator computes distribution over $B$. 
	    \State Let $l_j$ be the probability the generator assigns to output $j$
	    \State Generator draws with repetition discrete samples $b_j$ from the above distribution.
	    \State The semantic engine based discriminator estimates the probability $p_j$ of $b_j$ being real
	    \State Let $batch \_ mean \_ score$ be the average $p_j$
        \State Update generator parameters $\theta$ with a gradient step: $\nabla J(\theta) = p_j \times \nabla_{\theta}(\log(l_j))$.  
    \EndWhile
    \EndWhile
\end{algorithmic}
\caption{Basic SPGAN (simplified)}
\label{alg:gan1}
\end{algorithm}

\subsection{Comments on sequence generation}\label{sec:semSeqGen}
In order to scale the basic idea to sequences, we adjust the reward function defined in Section~\ref{sec:semanticGans} for sequences as in SeqGAN \cite{SeqGan}. More formally:
\begin{equation}
    J(\theta) = \sum_{t = 1}^{T} \sum_{j \in outputs} p(j^t|s_t, j_{1..t-1}) \times l(s_t, j_{1..t})
\end{equation}
\begin{equation}
    \nabla J(\theta) = \sum_{t = 1}^{T} \sum_{j \in outputs} p_j^t \times \nabla_{\theta}(\log(l_j^t))
\end{equation}

In contrast to \cite{SeqGan} we do not use random extensions of prefixes to reward prefixes of generated data, but instead we use the discriminator to reward prefixes by training it to distinguish between prefixes of real and generated sequences. In order to prevent mode collapse we use a simple technique where, instead of always generating complete sequences, we let the generator extrapolate prefixes of random length of randomly selected actual data for some fraction of the steps.

\subsection{Adaptation for continuous data}
The basic framework is presented for a discrete set of outputs. In order to apply it to continuous data we used the following: 

\begin{definition} \label{def:discretization} Let $B = 1 \ldots b$ denote the number of possible discrete outputs. A discretization schema is a pair of (potentially stochastic) functions, {\em projection} $\psi: \cal{X} \rightarrow B$, and {\em representative} $\phi: B \rightarrow \cal{X}$, Such that $\psi(\phi(\psi(x))) \sim \psi(x)$ for all $x \in \cal{X}$. The function $\phi(\psi(.))$ is called {\em reverse discretization}.
\end{definition}
For example if $\cal{X}$ represents geo-coordinates, it is possible to define a projection via a grid so each output $b$ represents a grid cell. A representative can be selected by choosing a random point over the corresponding cell.  We have found it highly useful to use reverse discretization for the actual data points. Intuitively, without doing this separation, it may be too easy for the engine (for example if some column is almost always an integer). We apply reverse discretization on the real data and the representative function on the generated data every time we produce training data for the semantic engine. This repeated randomization helps prevent over-fitting. The semantic engine works on the original domain $\cal X$ to allow it to use meaningful features from $\cal F$. With this schema, the key property \ref{def:keyProperty} is preserved in the sense that if the AUC is small then no feature separates the actual data (after reverse discretization) and the generated data. We omit the formalization for brevity.   We use two versions of the generator network architecture, one which views every discrete output in $1, \ldots, b$ separately and uses soft-max to choose among them, and one that replaces every discrete output by its reverse discretization to get better numeric generalization.
An alternative approach to discretization, that we did not implement in this work, is to generate parameters for some simple distributions and sample them to get the actual values.

\begin{figure*}[h]
\includegraphics[width=\textwidth]{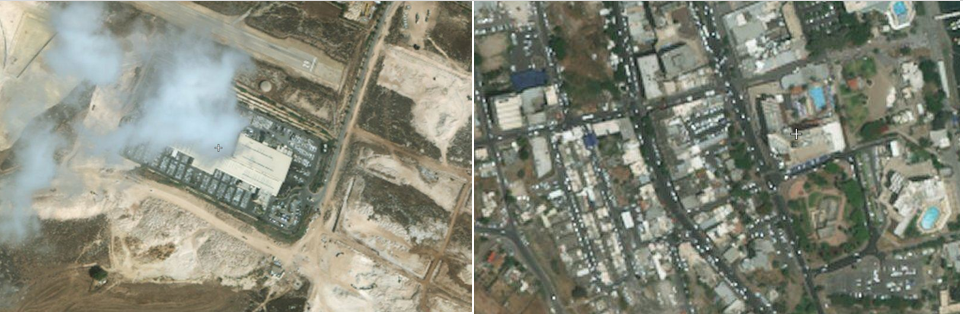}
\caption{Cellular antennae on buildings. Left: low ranked, Right: high ranked.}
\label{fig:anomaliesCellular}
\end{figure*}

\section{Experiments}
This section applies our method to three domains with very different characteristics. We start by studying the problem of generating locations of cellular antennae given a few hundreds sample points. This problem appears significantly different from problems to which GANs were typically applied. We continue with problems of molecule generation and demonstrate our method both in the case of rich and scarce data. Finally, we study an artificial problem where we can systematically measure the quality of the results and explore some properties of our method.

\noindent 
{\bf Settings:} With a single exception, all experiments took minutes to several hours on a standard laptop and used only simple generator architectures. The large molecule generation experiment required a single Tesla V100 GPU. The code is written in Tensorflow \cite{abadi2016tensorflow}. For the antennae experiment we utilized a commercial engine in order to get a rich hypothesis space, all other experiments used only open source dependencies.  All code, datasets, and intermediate byproducts can be found on our GitHub repository \footnote{https://github.com/SparkBeyond/public-research/tree/master/moleculesGeneration}.

\subsection{Generating locations of cellular antennas}
Consider applications in which the goal is to find good locations for new stores, oil wells, cellular antennae, etc. We know where such objects were built but we don't know which places where avoided. A natural approach for many such problems is to first construct a good single class model for the data at hand and then to rank potential candidates according to it. The challenge is to generate helpful negative points. To the best of our knowledge such problems were never studied in generative settings.

In this experiment we are given locations of cellular antennae which are placed on top of buildings in a certain country (latitude and longitude). We strive to produce a model that will both 'discover' properties of locations where such antennae are placed and will be able to apply this knowledge to other regions as well. Since we are given coordinates only, without semantic features coming from external data the ability to generalize is essentially nonexistent. The space of semantic features of the engine included a wide range of functions that extract information from OpenStreetMap data. The resulting space contained over 500,000 potential features - orders of magnitudes more than the number of points. 

We divided the country into three separate areas - north, middle, and south with some buffers in between. These regions differ from both geographic and socio-economic perspectives. The generator was trained over the middle part of the country that contained only about 850 points.

It is interesting to compare the {\em features} resulting from various iterations of the generator. Tables \ref{tab:cellularFeatures1} and \ref{tab:cellularFeatures5} show the top features from both the initial and the fifth generations of the experiment respectively. While the initial discriminator mainly indicate that antennas are placed in more populated areas, the later model shows a strong tendency towards commercial districts, shopping malls and the like. Note also that the radius of the selected features of the later discriminator are much smaller than the initial one. In other words, the model starts with coarse grained properties of the underlying data and then uncover finer semantic attributes. The information gain \cite{Quinlan:Trees} of the top feature was 0.67 in the initial generation. In the 6th and final generation none of the many potential features has an information gain of above 0.09.  This means that the generated points are semantically much closer to the actual points than the initial ones. The out of sample AUCs decreased from 0.97 to 0.71 in the 6th generation.  

\begin{table}[h]
\begin{tabular}{ |l|l| }
 \hline
 & Feature \\
 \hline
 1 & OSM items within 500 meters radius\\ 
 2 & Bus stop within 500 meters \\  
 3 & Known items within 2Km \\
 4 & ATM in less than 3Km \\  
 \hline
\end{tabular}
\caption {Top features of the initial generation}
\label{tab:cellularFeatures1}
\end{table}

\begin{table}[h]
\begin{tabular}{ |l|l| }
 \hline
 & Feature \\
 \hline
 1 & OSM item in less than 100 meters \\ 
 2 & Amenity in less than 100 meters \\  
 3 & Brand in less than 500 meters \\
 4 & A certain governmental site in less than 3Km \\  
 \hline
\end{tabular}
\caption {Top features of the fifth generations}
\label{tab:cellularFeatures5}
\end{table}

We now consider building a single class model in order to transfer the results to regions. We compare two single class classifiers. Both were trained with the actual locations in the middle region as well as equally sized negative data taken from the same bounding box. The first classifier, called {\em uniform}, was given random points as negatives. The second classifier, termed {\em semantic}, blended random points with generated samples. This way its negative set contains data with at least two levels of semantic proximity to the real data, facilitating both coarse and fine grain separation. 

Table \ref{tab:transferCellular} describes the performance of the discriminators described above when operated on the northern and southern regions of the country. In both cases, the test set contained the actual locations along with 10,000 random points. The semantic classifier is better on both regions. While both discriminators had good performance on the (sparsely populated) southern part, the semantic discriminator almost {\em halved} the error of the uniform discriminator on the (more difficult) northern part.

\begin{table}[h]
\centering
\begin{tabular}{ |l|l|l|l|}
\hline
Metric              & Region    & Semantic          & Uniform \\
\hline
Precision @ 50    & North     & 0.8               & 0.62  \\
Precision @ 100   & North     & 0.7               & 0.57  \\
\hline
Precision @ 50    & South     & 0.88                  & 0.86  \\
Precision @ 100   & South     & 0.82                  & 0.80  \\
\hline
\end{tabular}
\caption {Semantic and uniform discriminators transferred}
\label{tab:transferCellular}
\end{table}
The model gives rise to interesting anomalies. Figure \ref{fig:anomaliesCellular} shows aerial photographs of two antennae, the one that is ranked highest by the discriminator and one that is ranked the lowest. The differences between the two are evident from the pictures. The normal antenna resides in a commercial/touristic district. The abnormal one however is located in a remote industrial area.

\subsection{Generating molecules with desired properties} \label{sec:mol}
Generating new molecules and materials is an area of great interest recently \cite{gomez2018automatic,shahar,molecularNature}, trying to bring the power of machine learning to drive innovation in the physical world, speed up drug discovery, and other areas. We use this domain both to compare our work to state of the art and to explore effects of the sample size. Given a family of source molecules, our goal is to preserve key chemical properties.  Recently, \cite{GuacaMol} presented an evaluation framework, called GuacaMol, for molecule generation tasks along with a benchmark dataset. That paper showed near perfect results when basing on a huge family of general molecules (over a Million). However, expecting so many data points is non-realistic for most domains (and even for many applications in chemistry and pharmacology). We argue that domain knowledge, in the form of major properties to preserve, can improve the generator performance with limited data as the generalization capability of the discriminator increases. We show that the results of the best performing molecule generator from \cite{GuacaMol} sharply decline with the number of available data points. We then show that applying SPGAN significantly improves the results. 

\subsubsection{Experiment setup and baseline models}
In all our experiments the molecules are represented in SMILES format \cite{SmilesRep}. The training dataset is derived from the ChEMBL database as sampled by GuacaMol. For comparison we consider two methods. First, the best performing method from \cite{GuacaMol} (smilesLSTM), based on recurrent neural network and max likelihood (MLE) training over the smiles character generation. Second, is a sequence generator based on SeqGan \cite{SeqGan}. We used the evaluation metrics provided by GuacaMol. All generators copied the layered LSTM architecture from smilesLSM. 

For our SPGAN discriminator we use gradient boosted trees with relevant semantic features driven by domain knowledge. We first take molecule validity and all chemical properties whose KL-divergence is measured by GuacaMol\footnote{The KL divergence provided by GuacaMol is based on the usual KL-divergence but is not equivalent. In particular, higher value means better alignment between the source and generated distributions.}. These include: 'Weight', 'logP', 'BertzCT', 'TPSA', 'numHAcceptors', 'numHDonors', 'NumRotatableBonds', 'NumAliphaticRings', 'NumAromaticRings'. We excluded the ChemNet embedding which is a long vector with unclear semantics that we did not try to preserve. We also added character level counts and string length. 

We always start with MLE training(SMILES LSTM) until saturation of the negative log likelihood measured on a validation dataset of 5k molecules. For GAN methods we then continue training with a new goal. We use a test set of 10k molecules for the evaluation presented next. We repeat the experiments with training sizes of 1k, 10k, and 100k molecules. 

\subsubsection{Evaluation}
We use the code and metrics from GuacaMol for measuring the following metrics: Validity - percentage of generated molecules which are chemically valid, Uniqueness - percentage which do not repeat previously generated molecule, Novelty - percentage of unique molecules not appearing in training data, and KL divergence over an array of chemical properties. Note that Uniqueness and Novelty calculations are affected by validity values below 0.5 and 0.1 respectively. The KL metric is calculated only on the subset of unique valid molecules generated. 

Table \ref{tab:molecules} show the comparison between the three methods on various population sizes. SPGAN outperforms the other models with significantly more valid molecules while retaining the similarity between the distributions of the properties of the actual and generated data. This advantage grows as the data size shrinks. Interestingly, the results initially dropped when either SPGAN or SeqGan was applied after SMILES-LSTM and we did not manage to avoid that even by interleaved execution. We leave this for future study.

\begin{table*}[h]
  \centering
  \begin{tabular}{|c|c|c|c|c|c|}
        \hline
    Generator & Train size & Validity & Uniqueness & Novelty & KL divergence \\
    \hline
    SEQGAN & 1k & 0.03 & 0.29 & 0.06 & 0.62   \\
    \hline
    SMILES LSTM & 1k & 0.07 & 0.71 & 0.14 & 0.65    \\
    \hline
    SPGAN  & 1k & \textbf{0.35} & \textbf{0.97} & \textbf{0.68} & 0.63 \\
    \hline
    SEQGAN & 10k & 0.24 & .99 & .48 & 0.76    \\
    \hline
    SMILES LSTM & 10k & 0.42 & 1.0 & 0.84 & 0.78   \\    
    \hline
    SPGAN & 10k  & \textbf{0.57} & 0.99 & \textbf{.99} & 0.79\\
    \hline
    SEQGAN & 100k & 0.43 & 1.0 & 0.84 & 0.76  \\
    \hline
    SMILES LSTM & 100k  & 0.77 & 1.0 & 0.97 & 0.84    \\
    \hline
    SPGAN & 100k  & \textbf{0.87} & 1.0 & \textbf{0.99} & 0.83 \\
    \hline
  \end{tabular}
\caption{Molecule generation with different training sizes}
\label{tab:molecules}
\end{table*}

\subsubsection{Interpretation of the results}
Deep learning techniques are sensitive to the number of training rows. This explains the difference between the near perfect results in the GuacaMol metrics when a Million molecules were utilized for training and hardly producing any valid molecules when a thousand points were used. Complex patterns yet with limited data lead to over-fitting. By focusing on semantic properties we can not only highlight what we care about (validity, KL, AUC) but also add domain knowledge that boosts the generalization power of the discriminator. As Proposition \ref{def:keyProperty} states, this generalization transfers to the generator.

\subsection{Generating Lines from Small Data}

This experiment explores a simple artificial problem of generating straight lines of the form $y_t = \beta \cdot t + \epsilon$ where the slope $\beta$ is chosen uniformly in $[0,1]$ and $\epsilon$ is a small Gaussian noise. We introduce two modifications to it. We want to use as little training data as possible, and we also want to be able to extrapolate lines from given prefixes which are not part of the train. The first modification aims to incorporate the aspect of data scarcity. The second modification is aimed to both demonstrate the generalization power of the generator and to show that it can do a useful task beyond the basic GAN capability. The discretization is via 49 bins so the optimal multiplicative error is about 1\%. Note that in the experiment below there are less training points than bins forcing the model to have a good generalization power.

The experiment utilizes only a handful of semantic functions. The first function is the standard deviation of the difference sequence $f(y) = std(y_n -y_{n-1}, \ldots, y_1 - y_0)$. An invariant of the data is that the value of $f$ is small, governed only by the noise term. We also added features for the mean difference and the length. For the discriminator we use a standard scikit learn logistic regression classifier to avoid overfitting.  The topology of the generator is based on a simple LSTM network. 

\begin{figure}[h]\label{fig:lines}
    \includegraphics[width=0.5\textwidth,height=0.35\textwidth]{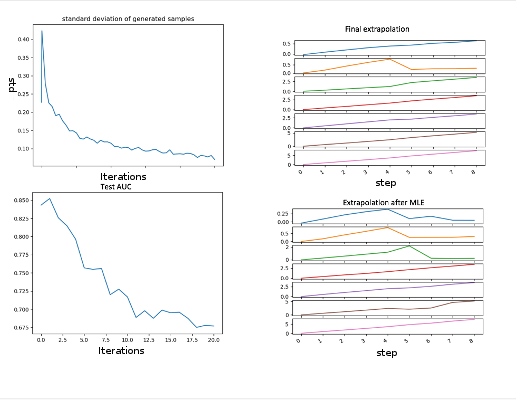}
  \caption{Extrapolating lines given only 16 training instances}
\end{figure}

In order to have a reasonable baseline we first use MLE training of the generator until saturation. This method attempts to maximize the likelihood that the LSTM assigns to the actual data and often provides surprisingly good results (e.g. \cite{SeqGan}). Figure \ref{fig:lines} describes the results of the experiment with only 16 training points. While MLE already produces reasonable results, the value of the invariant $f(.)$ on the generated samples is reduced  by the SPGAN iterations by a factor of 3 (from 0.22 to 0.07). This phenomenon was reproduced in many settings with the number of training points ranging from 8 to 200. The extrapolation task, while not perfect, is also significantly better than the MLE's. 

\vspace{.5cm}
\section{Conclusion}
This paper provides a generic method for constructing generators with indistinguishability guarantees over black box computable functions. Incorporating more prior knowledge in the shape of semantic functions also allows generalization with fewer training samples. Many open challenges stem from this work. Among them, the effect of the power of the discriminator (classifier topology, feature space, number of training examples) calls for further study. In addition, a single class semantic classifier can be used to filter the intermediate generator's results. This can potentially overcome inherent expressiveness limitations of the generator. Finally, the fact that semantic properties of data are revealed in a gradually finer semantic granularity has the potential to bring insight to many domains of interest. 
\vspace{.5cm}

\bibliographystyle{aaai}
\bibliography{semgans}

\end{document}